\pdfoutput=1

\documentclass[11pt]{article}

\usepackage[]{naacl2021}

\usepackage{times}
\usepackage{latexsym}

\usepackage[T1]{fontenc}

\usepackage[utf8]{inputenc}

\usepackage{microtype}

\usepackage{latexsym}

\usepackage{colortbl}

\usepackage{arydshln}
\usepackage{microtype}
\usepackage{booktabs}
\usepackage{multirow}
\usepackage{subfigure}
\usepackage{graphicx}
\usepackage{colortbl}
\usepackage{amsmath}
\usepackage{amssymb}
\usepackage{pifont}
\usepackage{enumitem}
\usepackage{algorithm}
\usepackage{algorithmic}
\usepackage{arydshln}

\title{\textit{G}\textit{Sum}: A {G}eneral Framework for \textit{G}uided Neural Abstractive \textit{S}ummarization}%

\author{Zi-Yi Dou, Pengfei Liu, Hiroaki Hayashi, Zhengbao Jiang, Graham Neubig\\
  Language Technologies Institute, Carnegie Mellon University \\
  {\tt \{zdou, pliu3, hiroakih, zhengbaj, gneubig\}@cs.cmu.edu}}

\date{}
\definecolor{mypink}{rgb}{0.99, 0.4, 0.7}

\def\showcomments{1}
\if\showcomments1

\newcommand{\zd}[1]{\textcolor{orange}{\bf\small [#1 --ZD]}}
\newcommand{\gn}[1]{\textcolor{magenta}{\bf\small [#1 --GN]}}

\newcommand{\hh}[1]{\textcolor{red}{\bf\small [#1 --HH]}}

\else
\newcommand{\gn}[1]{}
\newcommand{\zd}[1]{}
\newcommand{\hh}[1]{}
\fi

\begin{document}
\maketitle

\begin{abstract}
Neural abstractive summarization models are flexible and can produce coherent summaries, but they are sometimes unfaithful and can be difficult to control. While previous studies attempt to provide different types of guidance to control the output and increase faithfulness, it is not clear how these strategies compare and contrast to each other. In this paper, we propose a  general and extensible guided summarization framework (\textbf{GSum}) that can effectively take different kinds of external guidance as input, and we perform experiments across several different varieties. Experiments demonstrate that this model is effective, achieving state-of-the-art performance according to ROUGE on 4 popular summarization datasets when using highlighted sentences as guidance. In addition, we show that our guided model can generate more faithful summaries and demonstrate how different types of guidance generate qualitatively different summaries, lending a degree of controllability to the learned models.\footnote{Code is available at \url{https://github.com/neulab/guided_summarization}.}
\end{abstract}

\section{Introduction}
Modern techniques for text summarization generally can be categorized as either \emph{extractive} methods~\cite{nallapati2017summarunner,narayan2018ranking,zhou2018neural}, which identify the most suitable 
words or sentences from the input document and concatenate them to form a summary, or \emph{abstractive} methods~\cite{rush2015neural,chopra2016abstractive,nallapati2016abstractive,paulus2018deep}, which generate summaries freely and are able to produce novel words and sentences.
Compared with extractive algorithms, abstractive algorithms are more flexible, making them more likely to produce fluent and coherent summaries. 
However, the unconstrained nature of abstractive summarization can also result in problems.
First, it can result in \emph{unfaithful} summaries~\cite{kryscinski2019evaluating}, containing factual errors as well as hallucinated content.
Second, it can be difficult to \emph{control} the content of summaries; it is hard to pick in advance which aspects of the original content an abstractive system may touch upon.
To address the issues, we propose methods for \emph{guided} neural abstractive summarization: methods that provide various types of guidance signals that 1) constrain the summary so that the output content will deviate less from the source document; 2) allow for controllability through provision of user-specified inputs.

\begin{figure}[t]
\centering
\includegraphics[width=0.45\textwidth]{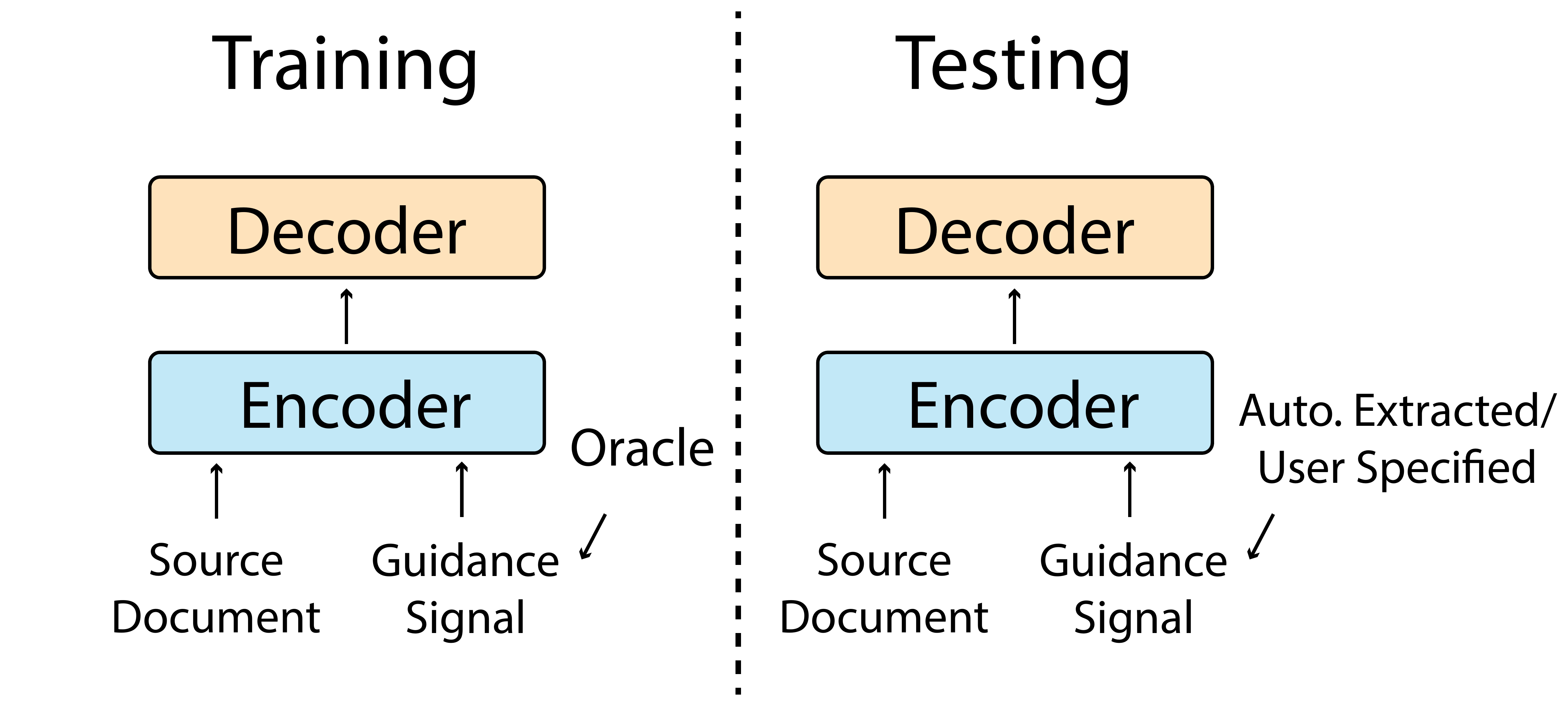}
\caption{Our framework generates summaries using both the source document and separate guidance signals. We use an oracle to select guidance during training and use automatically extracted or user-specified guidance at test time.}
\label{fig:fig1}
\end{figure}

\begin{table*}[t]
  \centering
  \small
  \begin{tabular}{@{}lllll@{}}
    \toprule
     \multirow{2}[4]{*}{\textbf{Work}} & \multicolumn{4}{c}{\textbf{Guidance Form}}  \\
\cmidrule{2-5}          & \textbf{Tokens} & \textbf{Triples} & \textbf{Sentences} & \textbf{Summaries} \\
    \midrule
    \citet{kikuchi-etal-2016-controlling}  & \ding{51}  (length tokens) & \ding{55} &  \ding{55} & \ding{55}    \\
    \citet{cao2018retrieve}   &  \ding{55}      &  \ding{55} &  \ding{55} &   \ding{51} (retrieved sums.)       \\
       \citet{li2018guiding}   &  \ding{51} (keywords)     &  \ding{55}   &  \ding{55} &  \ding{55}   \\
       \citet{liu2018generating}   &   \ding{55} &  \ding{55}  &   \ding{51} (highlighted sents.)    &  \ding{55}   \\
       \citet{liu2018controlling}   &   \ding{51} (length tokens) &  \ding{55}  &   \ding{55}    &  \ding{55}   \\
       \citet{fan2018controllable}  &  \ding{51}  (length, entity, style tokens)  &  \ding{55}  &  \ding{55} &  \ding{55}     \\
    \citet{zhu2020boosting}   &  \ding{55} &  \ding{51}  (relations)    &  \ding{55}  &  \ding{55}     \\
    \citet{jinsemsum}   &  \ding{55}       &  \ding{51} (relations)     &  \ding{55} &  \ding{55}    \\
    \citet{saito2020abstractive} &  \ding{51}  (keywords) &  \ding{55} &  \ding{51} (highlighted sents.) &  \ding{55} \\
    \midrule
    Ours   &   \ding{51} (keywords) & \ding{51} (relations) & \ding{51} (highlighted sents.) & \ding{51} (retrieved sums.) \\

    \bottomrule
    \end{tabular}%
  \caption{A comparison of different guided neural abstractive summarization models. Previous works have tried to provide guidance in different forms, including tokens, triples, sentences and summaries. Our proposed framework can incorporate them together and we have experimented with all four forms.}
  \label{tab:relatedwork}%
\end{table*}%

There have been some previous methods for guiding neural abstractive summarization models. For example,~\citet{kikuchi-etal-2016-controlling} specify the length of abstractive summaries,~\citet{li2018guiding} provide models with keywords to prevent the model from missing key information, and ~\citet{cao2018retrieve} propose models that retrieve and reference relevant summaries from the training set. 
While these methods have demonstrated improvements in summarization quality and controllability, each focuses on one particular type of guidance -- it remains unclear which is better and whether they are complementary to each other. 
In this paper, we propose a \emph{general and extensible guided summarization framework} that can take different kinds of external guidance as input.
Like most recent summarization models, our model is based on neural encoder-decoders, instantiated with contextualized pretrained language models, including BERT~\cite{Devlin2019BERTPO} and BART~\cite{Lewis2019BARTDS}.
With this as a strong starting point, we make modifications allowing the model to attend to \emph{both} the source documents and the guidance signals when generating outputs.
As shown in Figure~\ref{fig:fig1}, we can provide automatically extracted or user-specified guidance to the model during test time to constrain the model output.
At training time, to encourage the model to pay close attention to the guidance, 
we propose to use an \emph{oracle} to select informative guidance signals -- a simple modification that nonetheless proved essential in effective learning of our guided summarization models. 
Using this framework, we investigate four types of guidance signals: (1) highlighted sentences in the source document, (2) keywords, (3) salient relational triples in the form of (subject, relation, object), and (4) retrieved summaries.

We evaluate our methods on 6 popular summarization benchmarks.
Our best model, using highlighted sentences as guidance, can achieve state-of-the-art performance on 4 out of the 6 datasets, including 1.28/0.79/1.13 ROUGE-1/2/L improvements over previous state-of-the-art model on the widely-used CNN/DM dataset.
In addition, we perform in-depth analyses of different guidance signals and demonstrate that they are complementary to each other in that there is potential to aggregate their outputs together and obtain further improvements.
An analysis of the results also reveals that our guided models can generate more faithful summaries and more novel words.
Finally, we demonstrate that we can control the output by providing user-specified guidance signals, with different provided signals resulting in qualitatively different summaries.


\section{Background and Related Work}



\paragraph{Neural abstractive summarization}
typically takes a source document $\mathbf{x}$ consisting of multiple sentences $x_1, \cdots, x_{|\mathbf{x}|}$, runs them through an encoder to generate representations, and passes them to a decoder that outputs the summary $\mathbf{y}$ one target word at a time. Model parameters $\theta$ are trained to maximize the conditional likelihood of the outputs in a parallel training corpus $\langle \mathcal{X}, \mathcal{Y} \rangle$:
$$
\small
    \arg\max_{\theta} \sum_{\langle \mathbf{x}^i, \mathbf{y}^i \rangle \in \langle\mathcal{X}, \mathcal{Y}\rangle} \log p (\mathbf{y}^i \,|\, \mathbf{x}^i; \theta).
$$
Several techniques have been proposed to improve the model architecture. For example, models of copying~\citep{gu2016incorporating,see2017get,gehrmann2018bottom} allow words to be copied directly from the input to the output, and models of coverage discourage the model from generating repetitive words~\cite{see2017get}.


\paragraph{Guidance}
can be defined as some variety of signal $\mathbf{g}$ that is fed into the model in addition to the source document $\mathbf{x}$:
$$
\small
\arg\max_{\theta} \sum_{\langle \mathbf{x}^i, \mathbf{y}^i, \mathbf{g}^i \rangle \in \langle \mathcal{X}, \mathcal{Y}, \mathcal{G} \rangle} \log p (\mathbf{y}^i \,|\, \mathbf{x}^i, \mathbf{g}^i; \theta).
$$
Within this overall framework, the types of information that go into $\mathbf{g}$ and the method for incorporating this information into the model may vary. While there are early attempts at non-neural guided models~\cite{owczarzak2010overview,genest-lapalme-2012-fully}, here we focus on neural approaches and summarize recent work in Table~\ref{tab:relatedwork}.
For example,~\citet{li2018guiding} first generate a set of \emph{keywords}, which are then incorporated into the generation process by an attention mechanism.
~\citet{cao2018retrieve} propose to search the training corpus and \emph{retrieve} datapoint $\langle \mathbf{x}^j, \mathbf{y}^j \rangle$ whose input document $\mathbf{x}^j$ is most relevant to the current input $\mathbf{x}$, and treat $\mathbf{y}^j$ as a candidate template to guide the summarization process. Besides,
~\citet{jinsemsum} and ~\citet{zhu2020boosting} extract \emph{relational triples} in the form of (subject, relation, object) from source documents and represent them by graph neural networks.
The decoders then attend to the extracted relations to generate faithful summaries. A concurrent work by~\citet{saito2020abstractive} propose to extract \emph{keywords} or \emph{highlighted sentences} using saliency models and feed them to summarization models.

There are also works on controlling the summary length~\citep{kikuchi-etal-2016-controlling,liu2018controlling} and styles~\citep{fan2018controllable} by explicitly feeding the desired features to the model.
In addition,~\citet{liu2018generating} and~\citet{chen2018fast} follow a two-stage paradigm, in which a subset of the source document $\{ x_{i_1}, \cdots, x_{i_n} \}$ will first be selected by a pretrained extractor as \emph{highlighted sentences} and then be fed into the model encoder in the second stage with the rest of the text discarded.
\begin{figure}[t]
\centering
\includegraphics[width=0.48\textwidth]{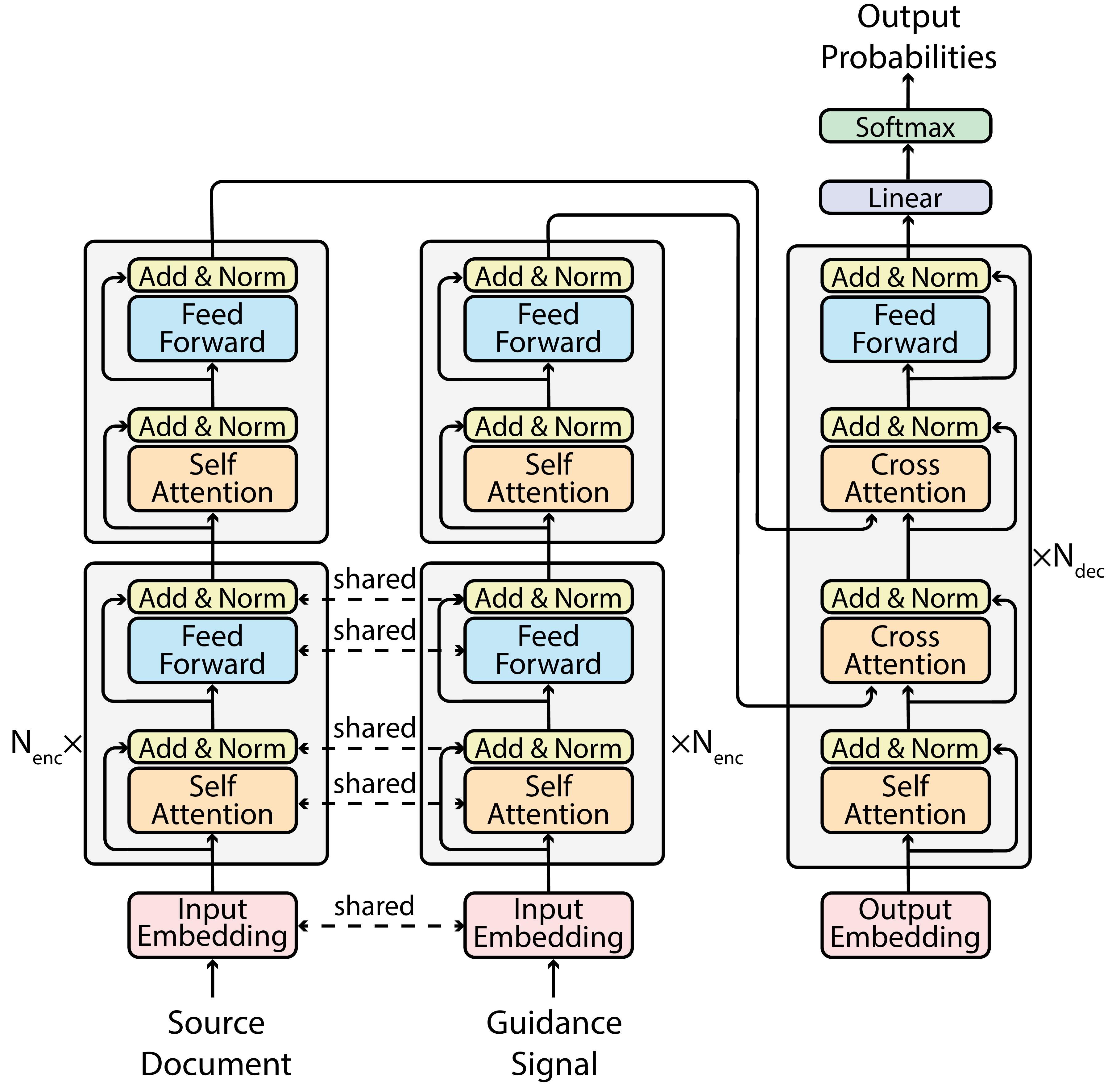}
\caption{General framework of our model. The two encoders encode the source document and guidance signal, which are attended to by the decoder.}
\label{fig:arch}
\end{figure}

\section{Methods}


Figure~\ref{fig:arch} illustrates the general framework of our proposed method. We feed both the source documents and various types of guidance signals to the model.
Specifically, we experiment with guidance signals including highlighted sentences, keywords, relations, and retrieved summaries, although the framework is general and could be expanded to other varieties of guidance as well.

\subsection{Model Architecture}
We adopt the Transformer model~\citep{Vaswani2017AttentionIA} as our backbone architecture, instantiated with BERT or BART, which can be separated into the encoder and decoder components.

\subsubsection{Encoder}
Our model has two encoders, encoding the input source document and guidance signals respectively.

Similar to the Transformer model, each of our encoders is composed of $N_{enc} + 1$ layers, with each encoding layer containing both a self-attention block and a feed-forward block:
$$
\small
\begin{aligned}
\mathbf{x} &= \textsc{LN}(\mathbf{x} + \textsc{SelfAttn}(\mathbf{x})),\\
\mathbf{x} &= \textsc{LN}(\mathbf{x} + \textsc{FeedForward}(\mathbf{x})),
\end{aligned}
$$
where $\small \textsc{LN}$ denotes layer normalization.
Note the source document and guidance signal do not interact with each other during encoding.

We share the parameters of the bottom $N_{enc}$ layers and the word embedding layers between the two encoders, because 1) this can reduce the computation and memory requirements; 2) we conjecture that the differences between source documents and guidance signals should be high-level, which are captured at top layers of the encoders.

\subsubsection{Decoder}
Different from the standard Transformer, our decoder has to attend to both the source document and guidance signal instead of just one input.

Concretely, our decoder is composed of $N_{dec}$ identical layers, with each layer containing four blocks. After the self-attention block, the decoder will first attend to the guidance signals and generate the corresponding representations, and hence the guidance signal will inform the decoder which part of the source documents should be focused on. Then, the decoder will attend to the whole source document based on the guidance-aware representations. Finally, the output representation will be fed into the feed-forward block:
$$
\small
\begin{aligned}
\mathbf{y} &= \textsc{LN}(\mathbf{y} + \textsc{SelfAttn}(\mathbf{y})),\\
\mathbf{y} &= \textsc{LN}(\mathbf{y} + \textsc{CrossAttn}(\mathbf{y}, \mathbf{g})),\\
\mathbf{y} &= \textsc{LN}(\mathbf{y} + \textsc{CrossAttn}(\mathbf{y}, \mathbf{x})),\\
\mathbf{y} &= \textsc{LN}(\mathbf{y} + \textsc{FeedForward}(\mathbf{y})).
\end{aligned}
$$

Ideally, the second cross-attention block allows the model to fill in the details of the input guidance signal, such as finding the name of an entity by searching through co-reference chains.

\subsection{Choices of Guidance Signals}


Before delving into the specifics of the types of guidance signal we used, 
we first note an important detail in training our model.
At test time, there are two ways we can define the guidance signal: 1) {\bf manual definition} where an interested user defines the guidance signal $\mathbf{g}$ by hand, and 2) {\bf automatic prediction} where an automated system is used to infer the guidance signal $\mathbf{g}$ from input $\mathbf{x}$.
We demonstrate results for both in experiments.

At training time, it is often prohibitively expensive to obtain manual guidance. Hence, we focus on two varieties of generating them: 1) \textbf{automatic prediction} using $\mathbf{x}$ as detailed above, and 2) \textbf{oracle extraction} where we use \emph{both} $\mathbf{x}$ and $\mathbf{y}$ to deduce a value $\mathbf{g}$ that is most likely useful in generating $\mathbf{y}$.

Theoretically, automatic prediction has the advantage of matching the training and testing conditions of a system that will also receive automatic predictions at test time.
However, as we will show in experiments, the use of oracle guidance has a large advantage of generating guidance signals that are highly informative, thus encouraging the model to pay more attention to them at test time.

With this in mind, we describe the four varieties of guidance signal we experiment with, along with their automatic and oracle extraction methods.


\begin{table*}[t]
\small
  \centering
  \begin{tabular}{@{}llrrrp{0pt}rrr@{}}
  \toprule
   \multirow{2}{*}{\bf Dataset}  & \multirow{2}{*}{\bf Source}  &
   \multicolumn{3}{c}{\bf \#Pairs } & &
   \multicolumn{2}{c}{\bf \#Tokens } &
   \multirow{2}{*}{\bf \#Ext} \\
  \cmidrule{3-5}\cmidrule{7-8}
 && Train& Valid & Test & & Doc. & Sum. & \\\midrule
 Reddit & Social Media & 41,675 & 645 & 645 & & 482.2 & 28.0 & 2 \\
 XSum & News & 203,028 & 11,273 & 11,332 & & 430.2 & 23.3 & 2 \\
 CNN/DM & News & 287,084 & 13,367 & 11,489 & & 766.1 & 58.2 & 3 \\
 WikiHow & Knowledge Base & 168,126 & 6,000 & 6,000 & & 580.8 & 62.6 & 4 \\
  NYT & News &  44,382 &  5,523 &  6,495 & & 1183.2 & 110.8 & 4\\
 PubMed & Scientific Paper & 83,233 & 4,946 & 5,025 & & 444.0 & 209.5 & 6 \\
 \bottomrule
    \end{tabular}
    \caption{ \label{tab:dataset} Statistics of the datasets. {\bf \#Ext} denotes the number of sentences we extract for extractive summarization.}
  \end{table*}

\paragraph{Highlighted Sentences.} The success of extractive approaches have demonstrated that we can extract a subset of sentences $\{ x_{i_1}, \cdots, x_{i_n} \}$ from the source document and concatenate them to form a summary. Inspired by this, we explicitly inform our model which subset of source sentences should be highlighted using extractive models.

We perform oracle extraction using a greedy search algorithm~\citep{nallapati2017summarunner,liu2019text} to find a set of sentences in the source document that have the highest ROUGE scores with the reference (detailed in Appendix) and treat these as our guidance $\mathbf{g}$.
At test time, we use pretrained extractive summarization models (BertExt~\cite{liu2019text} or MatchSum~\cite{zhong2020extractive} in our experiments) to perform automatic prediction. 

\paragraph{Keywords.} If we select full sentences, they may contain unnecessary information that does not occur in an actual summary, which could distract the model from focusing on the desired aspects of the input.
Therefore, we also try to feed our model with a set of individual keywords $\{w_1,\ldots,w_n\}$ from the source document.

For oracle extraction, we first use the greedy search algorithm mentioned above to select a subset of input sentences, then use TextRank~\citep{mihalcea2004textrank} to extract keywords from these sentences. We also filter the keywords that are not in the target summary. The remaining keywords are then fed to our models. 
For automatic prediction, we use another neural model (BertAbs~\cite{liu2019text} in the experiments) to predict the keywords in the target summary. 

\paragraph{Relations.} Relations are typically represented in the form of relational triples, with each triple containing a subject, a relation, and an object. For example, {\it Barack Obama was born in Hawaii} will create a triple {\it (Barack Obama, was born in, Hawaii)}.


For oracle extraction, we first use Stanford OpenIE~\citep{angeli2015leveraging} to extract relational triples from the source document. Similar to how we select highlighted sentences, we then greedily select a set of relations that have the highest ROUGE score with the reference, which are then flattened and treated as guidance.
For automatic prediction, we use another neural model (similarly, BertAbs) to predict the relation triples on the target side.

\paragraph{Retrieved Summaries.} Intuitively, gold summaries of similar documents with the input can provide a reference point to guide the summarization. Therefore, we also try to retrieve relevant summaries from the training data $\langle \mathcal{X}, \mathcal{Y} \rangle$.

For oracle extraction, we directly retrieve five datapoints $\{\langle \mathbf{x}_1, \mathbf{y}_1\rangle, \ldots, \langle \mathbf{x}_5, \mathbf{y}_5\rangle\}$ from training data whose summaries $\mathbf{y}_i$ are most similar to the target summary $\mathbf{y}$ using Elastic Search.\footnote{\url{https://github.com/elastic/elasticsearch}} 
For automatic prediction at test time, we retrieve five datapoints whose source documents $\mathbf{x}_i$ are most similar to each input source document $\mathbf{x}$ instead.

\section{Experiments}
\subsection{Datasets}
We experiment on 6 datasets (statistics in Table~\ref{tab:dataset}):
 \vspace{-15pt}
\paragraph{Reddit}\cite{kim2019abstractive} is a highly abstractive dataset and we use its {\it TIFU-long} version.
 \vspace{-6pt}
\paragraph{XSum}\cite{narayan2018don} is an abstractive dataset that contains one-sentence summaries of online articles from BBC.
 \vspace{-6pt}
\paragraph{CNN/DM}\cite{hermann2015teaching,nallapati2016abstractive} is a widely-used summarization dataset consisting of news articles and associated highlights as summaries. We use its non-anonymized version.
 \vspace{-6pt}
\paragraph{WikiHow}\cite{koupaee2018wikihow} is extracted from an online knowledge base and requires high level of abstraction.
 \vspace{-6pt}
\paragraph{New York Times (NYT)}\cite{sandhaus2008new} is a dataset that consists of news articles and their associated summaries.\footnote{\url{https://catalog.ldc.upenn.edu/LDC2008T19}}  We follow~\citet{kedzie2018content} to preprocess and split the dataset.
 \vspace{-6pt}
\paragraph{PubMed}\cite{cohan2018discourse} is relatively extractive and is collected from scientific papers.
 \vspace{-5pt}

\subsection{Baselines}
Our baselines include the following models:
 \vspace{-5pt}
\paragraph{BertExt}\cite{liu2019text} is an extractive model whose parameters are initialized with BERT~\cite{Devlin2019BERTPO}.
 \vspace{-5pt}
\paragraph{BertAbs}\cite{liu2019text} is an abstractive model with encoder initialized with BERT and trained with a different optimizer than its decoder.
 \vspace{-5pt}
\paragraph{MatchSum}\cite{zhong2020extractive} is an extractive model that reranks the candidate summaries produced by BertExt and achieves state-of-the-art extractive results on various summarization datasets.
 \vspace{-20pt}
\paragraph{BART}\cite{Lewis2019BARTDS} is an state-of-the-art abstractive summarization model pretrained with a denoising autoencoding objective.
 \vspace{-5pt}

\subsection{Implementation Details}
We build our models based on both BertAbs and BART, and follow their hyperparameter settings to train our summarizers. For our model built on BertAbs, there are 13 encoding layers, with the top layer randomly initialized and separately trained between the two encoders. For our model built on BART, there are 24 encoding layers, with the top layer initialized with pretrained parameters yet separately trained between the two encoders. The first cross-attention block of the decoder is randomly initialized whereas the second cross-attention block is initialized with pretrained parameters. BertAbs is used to predict guidance signals of relations and keywords during test time. Unless otherwise stated, we use oracle extractions at training time.

   \begin{table}[t]
       \small
  \centering
  \begin{tabular}{@{}lrrrr@{}}
  \toprule
   \bf Model &  \bf Guide & \bf R-1 &\bf R-2 &\bf R-L \\
  \midrule
  BertExt$^\ast$
  (Base) & - &43.25 & 20.24 & 39.63 \\
  BertAbs$^\ast$ & - & 41.72 & 19.39 & 38.76 \\
  BertAbs (Ours) & - & 41.58 & 18.99 & 38.56 \\
  \midrule
  \multicolumn{4}{l}{\it Ours} \\
  \midrule
    \multirow{2}{*}{BertAbs + Sentence} & Auto. & 43.78 & 20.66 & 40.66 \\
     & Oracle & 55.18 & 32.54 & 52.06 \\
    \addlinespace[0.2em]
    \hdashline
    \addlinespace[0.3em]
   \multirow{2}{*}{BertAbs + Keyword} & Auto. & 42.21 & 19.36 & 39.23 \\
    & Oracle & 45.08 & 22.22 & 42.07 \\
    \addlinespace[0.2em]
    \hdashline
    \addlinespace[0.3em]
    \multirow{2}{*}{BertAbs + Relation} & Auto. & 41.40& 18.66 & 38.40 \\
     & Oracle & 45.96 & 23.09 & 42.92\\
    \addlinespace[0.2em]
    \hdashline
    \addlinespace[0.3em]
    \multirow{2}{*}{BertAbs + Retrieve} & Auto. &40.88 & 18.24 & 37.99 \\
    & Oracle  & 43.69 & 20.53 & 40.71 \\ \bottomrule
    \end{tabular}
    \caption{ \label{tab:cnndm} Results (ROUGE; \citet{lin2004rouge}) on CNN/DM. ``Auto'' and ``oracle'' denote using automatically predicted and oracle-extracted guidance at test time respectively. Results with $^\ast$ are from \citet{liu2019text}.}
  \end{table}

  \begin{table}[t]
  \small
  \centering
  \begin{tabular}{@{}lrrr@{}}
  \toprule
   \bf Model & \bf R-1 &\bf R-2 &\bf R-L \\
   \midrule
  Oracle& 55.76 & 33.22 & 51.83 \\
  \midrule
  \multicolumn{4}{l}{\it Extractive} \\
  \midrule
  BertExt
  (Base)$^\ast$ & 43.25 & 20.24 & 39.63 \\
  BertExt (Large)$^\ast$ & 43.85 & 20.34 & 39.90 \\
  MatchSum$^\dagger$ & 44.41 & 20.86 & 40.55 \\
  \midrule
  \multicolumn{4}{l}{\it Abstractive} \\
  \midrule
  BertAbs$^\ast$ & 41.72 & 19.39 & 38.76 \\
    BertAbs (Ours) & 41.58 & 18.99 & 38.56 \\
  BertExtAbs$^\ast$ & 42.13 & 19.60 & 39.18 \\
  BART~$^\ddagger$ & 44.16 & 21.28 & 40.90 \\
  BART (Ours) & 44.66 & 21.53 & 41.35 \\
  \midrule
  \multicolumn{4}{l}{\it Ours} \\
  \midrule
    BertAbs + BertExt & 43.78 & 20.66 & 40.66 \\
    BART + MatchSum & \bf 45.94 & \bf 22.32 & \bf 42.48 \\
    \bottomrule
    \end{tabular}
    \caption{ \label{tab:cnndm_sota} Comparisons with state-of-the-art models on CNN/DM. The highest numbers are in {\bf bold}. Marked results are from  \citet{liu2019text}$^\ast$, \citet{zhong2020extractive}$^\dagger$, \citet{Lewis2019BARTDS}$^\ddagger$.}
  \end{table}

 \begin{table*}[t]
     \small
  \centering
    \setlength{\tabcolsep}{1.2pt}
  \begin{tabular}{@{}lrrrrrrrrrrrrrrrrrrrr@{}}
  \toprule
  \multirow{2}{*}{\bf Model} &\multicolumn{3}{c}{\bf Reddit} & &\multicolumn{3}{c}{\bf XSum} & &\multicolumn{3}{c}{\bf WikiHow} & &\multicolumn{3}{c}{\bf PubMed} & &\multicolumn{3}{c}{\bf NYT}  \\
   \cmidrule{2-4}\cmidrule{6-8}\cmidrule{10-12}\cmidrule{14-16}\cmidrule{18-20}
    & \bf R-1 &\bf R-2 &\bf R-L & & \bf R-1 &\bf R-2 &\bf R-L  & & \bf R-1 &\bf R-2 &\bf R-L & & \bf R-1 &\bf R-2 &\bf R-L & & \bf R-1 &\bf R-2 &\bf R-L \\
  \midrule
  Oracle&36.21&13.74 & 28.93 & &  29.79 & 8.81 & 22.66 & & 35.59 & 12.98 & 32.68 & & 45.12 & 20.33 & 40.19 & & 58.44 & 38.39 & 50.00\\
  \midrule
  \multicolumn{4}{l}{\it Extractive} \\
  \midrule
  BertExt (Base) & 23.86&5.85&19.11& & 22.86&4.48&17.16& & 30.40 & 8.67 & 28.32 & & 40.29 & 14.37 & 35.88 & & 45.98 & 25.29 & 42.46 \\
  MatchSum & 25.09& 6.17&20.13 & & 24.86&4.66&18.41 & & 31.85 & 8.98 & 29.58 & & 41.21 & 14.91 & 36.75 & & 46.98 & 26.67 & 43.62 \\
  \midrule
  \multicolumn{4}{l}{\it Bert-Based} \\
  \midrule
  BertAbs & \bf 26.92 & 6.35 & 19.81 & &  38.76 & \bf 16.33 & \bf 31.15 & & 38.16 & 15.06 & 34.71 & & 36.04 & 12.16 & 29.02 & & 49.94 & 31.44 & 46.67\\
  Ours (BertAbs + MatchSum) & 26.89 & \bf 6.75 & \bf  20.35 & & \bf 38.77 & 16.14 & 30.96 & & \bf 38.29 & \bf 15.10 & \bf 34.80 & & \bf 37.82 & \bf 12.32 & \bf 30.53 & &\bf  50.50 &\bf  31.57 &\bf  47.24\\
  \midrule
  \multicolumn{4}{l}{\it BART-Based} \\
  \midrule
   BART & \bf 35.00 & \bf 12.89 & \bf 27.96 & & \bf 45.51 & \bf 21.94 & \bf 36.75 & & 41.46 & \bf 17.80 & 39.89 & &  44.72 & 16.48 & 41.00 & & 54.13 & 35.15 & 47.00 \\
  Ours (BART + MatchSum) & 34.52 & 12.71 & 27.58 & & 45.40 & 21.89 & 36.67 & & \bf 41.74 & 17.73 & \bf 40.09 & & \bf 45.09 & \bf 16.72 & \bf 41.32 & & \bf 54.27 & \bf 35.37 & \bf 47.63  \\
  \bottomrule
    \end{tabular}
    \caption{ \label{tab:others} Results of our model guided with highlighted sentences on five datasets. Highest numbers in each section are in {\bf bold}. We use MatchSum to predict the guidance at test time. Extractive results are from \citet{zhong2020extractive}.}
  \end{table*}


\begin{table}[t]
       \small
  \centering
  \setlength{\tabcolsep}{2pt}
  \begin{tabular}{@{}rrrrrr@{}}
  \toprule
  \multicolumn{4}{c}{\bf Win [\%]} & & \multicolumn{1}{c}{\bf Combined}\\
  \multicolumn{1}{r}{\bf Sentence} & \multicolumn{1}{r}{\bf Keyword} &\multicolumn{1}{r}{\bf Relation} &\multicolumn{1}{r}{\bf Retrieve} & & \multicolumn{1}{c}{\bf R-1/R-2/R-L} \\ \midrule
  39.28 & 19.55&21.12&20.05& & 48.30/25.25/45.15 \\
 \bottomrule
    \end{tabular}
    \caption{ \label{tab:comp} No guidance signals can outperform all the other ones for all the test data, and aggregating the best outputs of the four guided models achieves significant improvements over the best single guided model (43.78/20.66/40.66 R-1/R-2/R-L scores).}
  \end{table}

\begin{table}[t]
       \small
  \centering
  \begin{tabular}{@{}lrrrr@{}}
  \toprule
   & \bf Sentence & \bf Keyword &\bf Relation &\bf Retrieve\\ \midrule
  \bf Sentence  & \underline{43.78}  & 46.11 & 46.20 & 46.27  \\
  \bf Keyword   & - & \underline{42.21} & 44.39 & 44.35 \\
  \bf Relation  & - & - & \underline{41.40} & 44.60\\
  \bf Retrieve  & - & - & - & \underline{40.88}\\
 \bottomrule
    \end{tabular}
    \caption{ \label{tab:comp2} Combining the best outputs of each pair of guidance signals leads to improvements (in terms of ROUGE-1), indicating every pair of guidance complements each other. The \underline{underlined} results are the model performance without combinations.} 
  \end{table}

\subsection{Main Results}
We first compare different kinds of guidance signals on the CNN/DM dataset using BertAbs, then evaluate the best guidance on the other five datasets using both BertAbs and BART.

\paragraph{Performance of Different Guidance Signals.} As shown in Table~\ref{tab:cnndm}, if we feed the model with automatically constructed signals, feeding either highlighted sentences or keywords can outperform the abstractive summarization baseline by a large margin. Especially, feeding highlighted sentences can outperform the best baseline by more than 1 ROUGE-L point. Using relations or retrieved summaries as guidance will not improve the baseline performance, likely because it is hard to predict these signals during test time.

If we use an oracle to select the guidance signals, all varieties of guidance can improve the baseline performance significantly, with the best-performing model achieving a ROUGE-1 score of 55.18. The results indicate that 1) the model performance has the potential to be further improved given a better guidance prediction model; 2) the model does learn to depend on the guidance signals.

\paragraph{Comparisons with State of the Art.} We then try to build our model on the state-of-the-art model, using highlighted sentences as guidance as it achieves the best performance on CNN/DM. First, we build our model on BART and train it with oracle-extracted highlighted sentences as guidance. Then, we use MatchSum to predict the guidance at test time. From Table~\ref{tab:cnndm_sota}, we can see that our model can achieve over 1 ROUGE-1/L point improvements compared with the state-of-the-art models, indicating the effectiveness of the proposed methods.

\paragraph{Performance on Other Datasets.} We report the performance of the highlighted sentence model on all the other five datasets in Table~\ref{tab:others}. Generally, the model works better when the dataset is more extractive. For abstractive datasets such as Reddit and XSum, our model cannot achieve performance increases when the abstractive summarization baseline is already rather strong.
For extractive datasets such as PubMed and NYT, on the other hand, our model can achieve some improvements over the baselines even though the abstractive baseline outperforms the extractive oracle model in some cases.
\subsection{Analysis}
We perform extensive analyses on CNN/DM to gain insights into our (BERT-based) models. Unless otherwise stated, we use oracle extractions at training time and automatic prediction at test time.

\begin{figure}[t]
\centering
\includegraphics[width=0.5\textwidth]{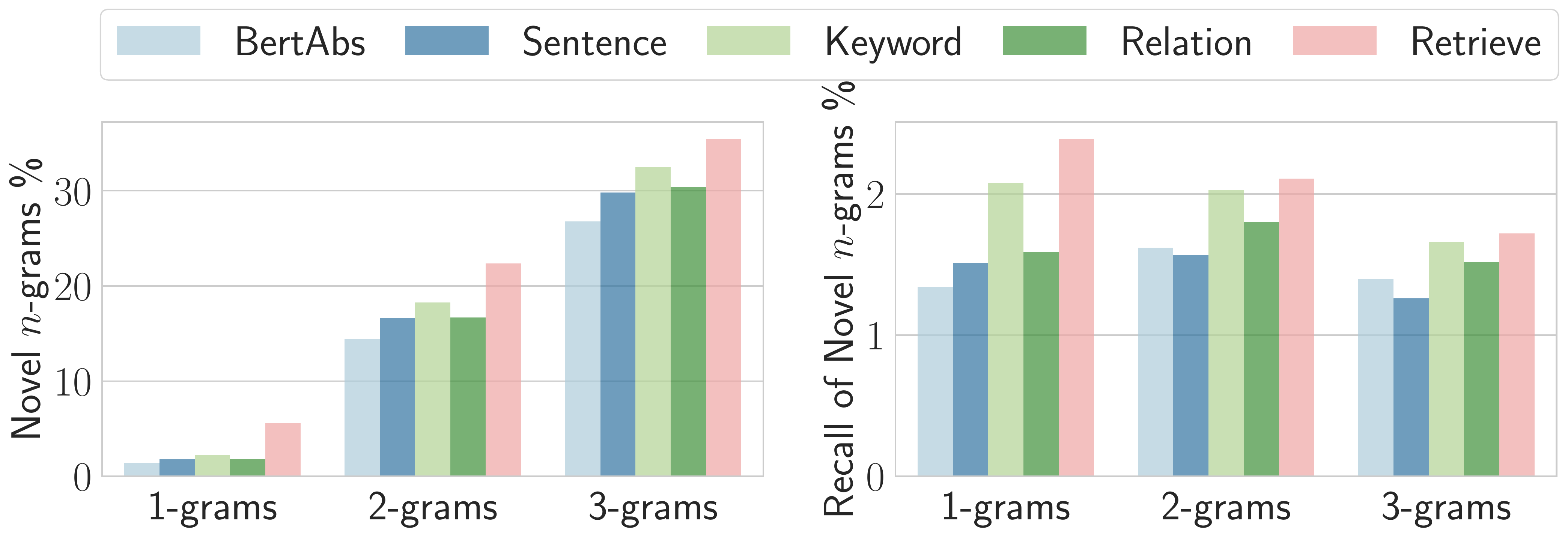}
\caption{Our model can generate more novel words and achieve higher recall of novel words in the gold reference compared with baseline.}
\label{fig:bar}
\end{figure}

\begin{table*}[tb]
    \scriptsize
    \centering
    \extrarowheight=\aboverulesep
    \addtolength{\extrarowheight}{\belowrulesep}
    \aboverulesep=0pt
    \belowrulesep=0pt
    \begin{tabular}{@{}c p{0.46\textwidth}  p{0.40\textwidth}}
     \toprule
   \multicolumn{1}{c}{ \bf Model} &  \multicolumn{1}{c}{ \bf Guidance} &  \multicolumn{1}{c}{\bf Output} \\\midrule
      \multicolumn{1}{c}{\cellcolor{gray!25} \bf Ref.} &  -  & nathan dailo has found a way to get his son to sleep in 42 seconds. in a youtube video he demonstrates how stroking his 3-month-old son's face with a white piece of tissue paper sends him to sleep. the video has received almost 26,000 views in just two weeks. \\\midrule
        \rowcolor{gray!25}
  \multicolumn{1}{c}{  \multirow{2}{*}{\cellcolor{gray!25} \bf Sentence}} & {\bf nathan dailo, from sydney, uploaded a video to his youtube channel demonstrating how he gets his three-month-old son seth to} drift off {\bf in just 42 seconds}. {\bf the clip that has now received almost 26,000 views} sees the father gliding a piece of white tissue paper over his son's face repeatedly until he nods off. in his youtube video, nathan dailo shows that {\bf by running tissue paper over his son seth makes him sleep.} & \textbf{nathan dailo, from sydney, \textit{uploaded a video to his youtube channel demonstrating how he gets his three-month-old son seth to}} \textit{sleep} {\bf in just 42 seconds}. {\bf the clip has now received almost 26,000 views}. \textit{\textbf{by running tissue paper over his son seth makes him sleep}}.  \\
   \multicolumn{1}{c}{ \cellcolor{gray!25} }& {\bf the clip that has now received almost 26,000 views} sees the father gliding a piece of white tissue paper over his son's face repeatedly until he nods off. in his youtube video, {\bf nathan dailo shows that by running tissue paper over his son seth makes him sleep}. {\bf one} grateful {\bf viewer} of the video {\bf commented:`this worked for my baby 5 times already so thanks!'} & {\bf nathan dailo}, from sydney, \textit{\textbf{shows that by running tissue paper over his son seth makes him sleep}} in 42 seconds. {\bf the clip has now received almost 26,000 views}. \textit{\textbf{one viewer commented:`this worked for my baby 5 times already so thanks!'}} \\\midrule
   \rowcolor{gray!25}
    \multicolumn{1}{c}{ \multirow{2}{*}{\bf Keyword} \cellcolor{gray!25}}& {\bf video} {\bf sleep} {\bf dailo} {\bf youtube} way {\bf seconds} {\bf demonstrates} {\bf weeks} piece {\bf received} & nathan {\bf dailo}, from sydney, uploaded a {\bf video} to his {\bf youtube} channel. he {\bf demonstrates} how he gets his three-month-old son seth to {\bf sleep} in just 42 {\bf seconds}. \textit{the clip has now \textbf{received} almost 26,000 views since being uploaded two \textbf{weeks} ago.} \\
    \multicolumn{1}{c}{ \cellcolor{gray!25}} & {\bf son} {\bf paper} sends found {\bf demonstrates} way views {\bf face} {\bf dailo} {\bf piece} & nathan {\bf dailo}, from sydney, posted a video to his youtube channel. he {\bf demonstrates} how he gets his three-month-old {\bf son} seth to sleep in just 42 seconds. \textit{the father glides a piece of white \textbf{tissue paper} over his son's \textbf{face} repeatedly until he nods off}. \\\midrule
   \multicolumn{1}{c}{ \cellcolor{gray!25} \multirow{2}{*}{\bf Relation}} & \cellcolor{gray!25} ({\bf father}, gliding, piece of {\bf tissue paper}), ({\bf clip}, sees {\bf father over}, {\bf his son}'s face) & \cellcolor{gray!25} \textit{a \textbf{father} from sydney has worked out a way of getting his three-month-old son to sleep}. in his youtube video, nathan dailo shows that by running {\bf tissue paper} over {\bf his son} seth makes him sleep in just 42 seconds. the {\bf clip} has now received almost 26,000 views since being uploaded two weeks ago. \\
    \multicolumn{1}{c}{\cellcolor{gray!25} }& ({\bf nathan dailo}, {\bf uploaded}, {\bf video to his youtube channel}), (has drifted, {\bf has received}, {\bf 26,000 views}) & \textit{\textbf{nathan dailo}, from sydney, \textbf{uploaded a video to his youtube channel}}. the clip {\bf has received 26,000 views} since being uploaded two weeks ago. it shows the father gliding a piece of white tissue paper over his son's face repeatedly until he nodded off.\\\midrule
    \multicolumn{1}{c}{ \cellcolor{gray!25} \multirow{2}{*}{\bf Retrieve}} & \cellcolor{gray!25} phil smith {\bf posted} the {\bf video} of him. {\bf his son} noah's tooth {\bf to his youtube channel} which went viral with over 100,000 {\bf views}. smith tied noah's tooth to dental attached to a golf ball. smith hits the ball and the tooth is gone {\bf in just seconds}, made by the expression in noah's face. & \cellcolor{gray!25} \textit{nathan dailo, from sydney, \textbf{posted} a \textbf{video to his youtube channel}}. he demonstrates how he gets {\bf his} three-month-old {\bf son} to sleep {\bf in just} 42 {\bf seconds}. the clip has now received 26,000 {\bf views} since being uploaded two weeks ago. \\
   \multicolumn{1}{c}{\cellcolor{gray!25}}  & karen wiltshire, 46, sends the babies to sleep by their and playing them the sound of a. it can take anything from a few minutes to half an hour for the tiny newborns to off to sleep. {\bf mother-of-two} from poole, dorset, {\bf has won an award in studio children photography for her work}. & \textit{\textbf{mother-of-two} from sydney, \textbf{has won an award in studio children photography for her work}}. in a youtube video, nathan dailo shows how he gets his three-month-old son seth to sleep in just 42 seconds. the clip has now received 26,000 views since being uploaded two weeks ago. \\
  \bottomrule
\end{tabular}
\vspace{-2mm}
\caption{Different guidance signals lead to qualitatively distinct outputs. Tokens that appear in both the guidance and output are {\bf bolded}. Sentences generated differently because of the effect of guidance content are {\it italicized}.}
\vspace{-2mm}
\label{tab:example}
\end{table*}

\paragraph{Novel $n$-grams.} While we sometimes provide information extracted from the source document as guidance signals, it is unclear whether the model will over-fit to and regurgitate this guidance, or still generate novel expressions. To measure this, we count the number of novel $n$-grams in the output summaries, namely $n$-grams that do not appear in the source document. As shown in Figure~\ref{fig:bar}, all of our guided models in fact generate \emph{more} novel $n$-grams than the baseline, likely because at training time the model is trained to 
compress and paraphrase the extracted information from the source document into the gold summary. In addition, our models cover more novel $n$-grams that are in the gold reference than baseline. The results indicate that our guided models can indeed generate novel expressions, and are not referencing the input guidance too strongly.

\paragraph{Complementarity of Different Guidance Signals.} 
While some guidance signals achieve worse performance than others, it is still possible to aggregate their outputs and obtain better performance if their outputs are diverse and they complement each-other. To verify this hypothesis, we try to select the best output of the four guidance signals for each test datapoint and investigate if we can aggregate their best outputs and achieve better performance.

Concretely, for each test input, we perform an oracle experiment where we compute the ROUGE score of each output of the four guidance signals and pick the best one. As shown in Table~\ref{tab:comp}, despite the fact that the highlighted sentence signal achieves the best overall performance, it still underperforms one of the other three varieties of guidance more than 60\% of the time. In addition, by aggregating their best outputs together, we can achieve a ROUGE-1/L point of 48.30/45.15, which significantly outperforms any single guided model. 
Further, we try to aggregate these guidance signals in a pairwise manner, and Table~\ref{tab:comp2} demonstrates that each guidance signal is complementary to each other to some extent. 
Thus, we can safely conclude that each type of guidance signal has its own merits and one promising direction is to utilize a system combination method such as \citet{hong-etal-2015-system} to aggregate the results together.

\paragraph{Controllability.}


It is also of interest what effect this guidance has on the model outputs qualitatively.
We sample several generated outputs (Table~\ref{tab:example}) and find that different provided signals can result in different outputs. Especially, for our sentence-guided model, providing the model with {\it by running tissue paper over his son seth makes him sleep} enables the model to generate the exact same sentence, and when the model is fed with {\it one grateful viewer of the video commented...}, it will generate {\it one viewer commented...}. The examples demonstrate that our model can generate summaries mostly faithful to the guidance signals while also performing abstraction.

\begin{table}[tb]
  \centering
  \small
  \begin{tabular}{@{}rrrrr@{}}
  \toprule
  \multirow{2}{*}{ \bf BertAbs} &   \multicolumn{4}{c}{\bf Ours} \\\cmidrule{2-5}
   &  \bf Sentence & \bf Keyword &\bf Relation &\bf Retrieve\\ \midrule
  2.117 & 2.393$^\ast$ & 2.347$^\ast$ & 2.303$^\ast$ & 2.310$^\ast$ \\
 \bottomrule
    \end{tabular}
    \vspace{-2mm}
    \caption{ \label{tab:faith} Human evaluation of the faithfulness of different model outputs. $^\ast$ indicates significant improvements ($p < 0.001$) over baseline with using bootstrap.}
    \vspace{-2mm}
  \end{table}

\paragraph{Faithfulness of Generated Summaries.} We also evaluate whether our generated summaries are faithful to the source document. We randomly sample 100 datapoints from the test set and ask 3 people from Amazon Mechanical Turk to evaluate their factual correctness. Each person gives a score between 1 and 3, with 3 being perfectly faithful to the source document. Table~\ref{tab:faith} shows that our guided model can generate more faithful summaries compared with the baseline.


\paragraph{Necessity of Using Oracles During Training.} As mentioned previously, we use an oracle to select guidance signals during training. In this part, we investigate if we can provide automatically constructed guidance to the model during training as well. Table~\ref{tab:oracle} shows that this methodology will lead to significantly worse performance. We conjecture that this is because when the relevancy between guidance and reference is weakened, the model will not learn to depend on the guidance signals and thus the model will be reduced to the original abstractive summarization baseline.
 \begin{table}[tb]
        \small
  \centering
  \begin{tabular}{@{}lrrrr@{}}
  \toprule
 \bf Train &  \bf Test & \bf R-1 &\bf R-2 &\bf R-L \\
  \midrule
   \multirow{2}{*}{Oracle} & Auto & 43.78 & 20.66 & 40.66 \\
     & Oracle & 55.18 & 32.54 & 52.06\\
   \midrule
   \multirow{2}{*}{Auto} & Auto & 41.61 & 19.04 & 38.65 \\
     & Oracle & 43.07 & 20.79 & 40.13\\
\bottomrule
    \end{tabular}
    \vspace{-2mm}
    \caption{ \label{tab:oracle} Using automatically constructed guidance during training degrades the performance significantly.}
    \vspace{-3mm}
  \end{table}

\section{Conclusion}
\vspace{-7pt}
We propose a general framework for guided neural summarization, using which we investigate four types of guidance signals and achieve state-of-the-art performance on various popular datasets.
We demonstrate the complementarity of the four guidance signals, and find that our models can generate more novel words and more faithful summaries. We also show that we can control the output by providing user-specified guidance signals.

Given the generality of our framework, this opens the possibility for several future research directions including 1) developing strategies to ensemble models under different guidance signals; 2) incorporating sophisticated techniques such as copy or coverage over the source document, the guidance signal, or both; and 3) experimenting with other kinds of guidance signals such as salient elementary discourse units.

\section*{Acknowledgements}

We thank Shruti Rijhwani, Yiran Chen, Jiacheng Xu and anonymous reviewers for valuable feedback and helpful suggestions.
This work was supported in part by a grant under the Northrop Grumman SOTERIA project and the Air Force Research Laboratory under agreement number FA8750-19-2-0200. The U.S. Government
is authorized to reproduce and distribute reprints for Governmental
purposes notwithstanding any copyright notation thereon. The views and
conclusions contained herein are those of the authors and should not be
interpreted as necessarily representing the official policies or
endorsements, either expressed or implied, of the Air Force Research
Laboratory or the U.S. Government.

\bibliographystyle{acl_natbib}
\bibliography{naacl2021}
\appendix

\section{Greedy Selection Algorithm}
Algorithm \ref{alg:main} demonstrates how we use an oracle to select a subset of source sentences that have the highest ROUGE scores with the reference summary. We use a similar algorithm to select the relation triples as well. Concretely, we flatten each relational triple $(s, r, o)$ by concatenating its elements together and treat each concatenated text as a source sentence, then use Algorithm \ref{alg:main} to select the relation triples greedily.

 \begin{algorithm}[t]
\caption{\label{alg:main} Greedy Selection Algorithm}
\begin{algorithmic}
    \REQUIRE{A source document $\mathbf{x}$ consisting of multiple sentences $\{x_1, \cdots, x_{|\mathbf{x}|}\}$, its reference summary $\mathbf{y}$, and a pre-defined integer $N$}
    \ENSURE{Oracle-selected highlighted sentences $\mathbf{o}$}
    \STATE{$\mathbf{o} = \{\}$}
    \FOR{i = $1, \cdots, N$}
        \STATE{max\_rouge$=0$}
        \FOR{$\mathbf{s}$ in $\mathbf{x} / \mathbf{o}$}
            \STATE{rouge\_1, rouge\_2 = cal\_rouge($\mathbf{o} \cup \{\mathbf{s}\}$) }
            \STATE{cur\_rouge = rouge\_1 + rouge\_2}
            \IF{cur\_rouge $>$ max\_rouge}
                \STATE{max\_rouge $=$ cur\_rouge}
                \STATE{max\_sent $= \mathbf{s}$ }
            \ENDIF
        \ENDFOR
        \IF{max\_rouge $== 0$}
            \STATE{\bf break}
        \ENDIF
        \STATE{$\mathbf{o} = \mathbf{o}\cup \{$ max\_sent $\}$}
	\ENDFOR
	\RETURN{$\mathbf{o}$}
\end{algorithmic}
\end{algorithm}

\section{Analysis}
We perform more analysis on CNN/DM in this section. Unless otherwise stated, we use oracle extractions at training time and BertAbs as our base model.
\subsection{Controllability}

\begin{figure}[t]
\centering
\includegraphics[width=0.45\textwidth]{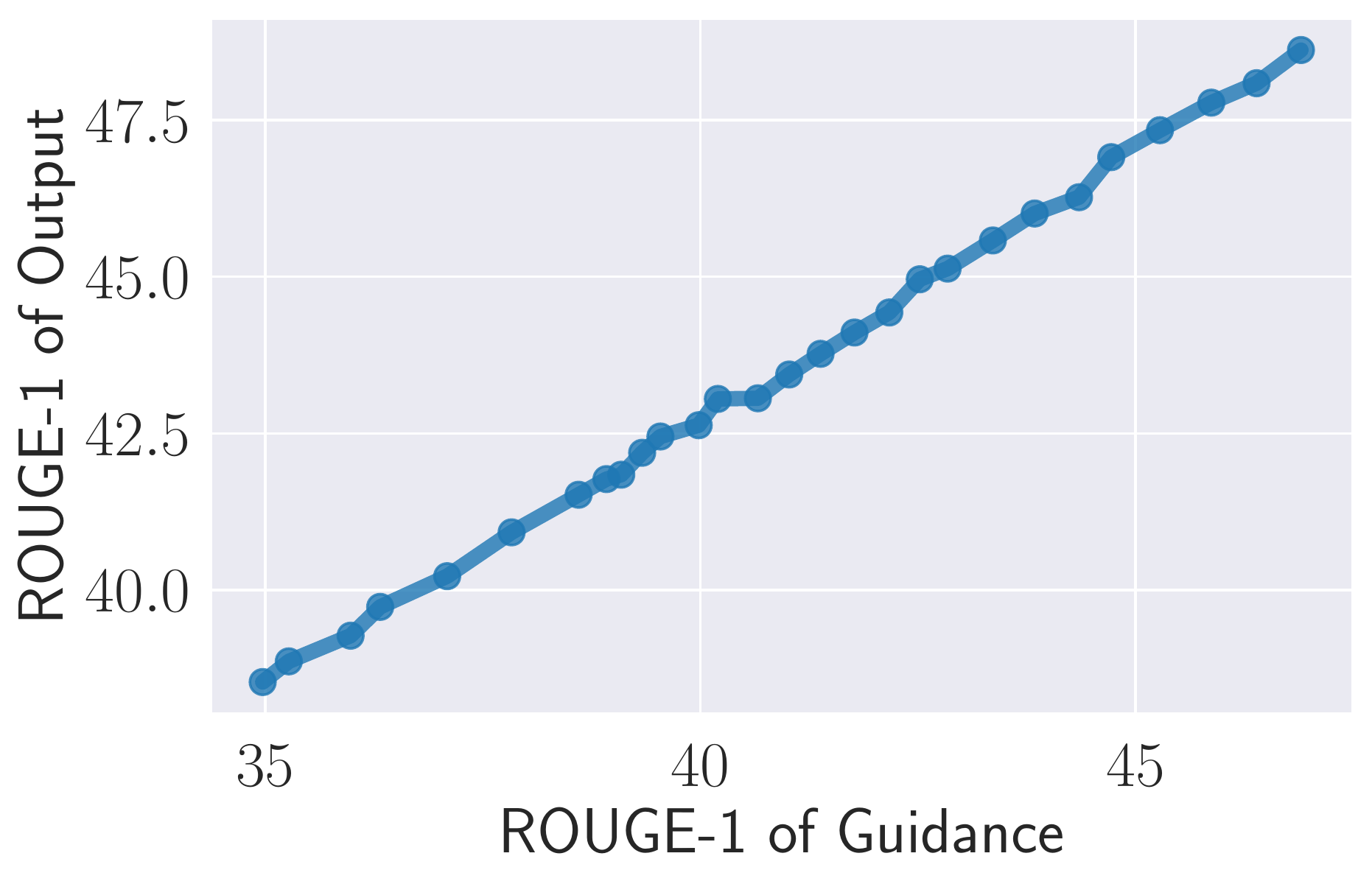}
\caption{There is a strong correlation between the guidance quality and output quality, demonstrating the controllability of our guided model. }
\label{fig:control}
\end{figure}

 \begin{table}[t]
       \small
  \centering
  \begin{tabular}{@{}lrrrrr@{}}
  \toprule
   \bf Model &  \bf Ref. & \bf Guidance & \bf R-1 &\bf R-2 &\bf R-L \\
  \midrule
    \multirow{4}{*}{Sentence} & \multirow{2}{*}{1st} &  1st & 49.49 & 29.39 & 46.25 \\
    &  &  2nd & 28.66 & 10.09 & 26.05 \\
    \addlinespace[0.2em]
     \cdashline{2-6}
     \addlinespace[0.2em]
    & \multirow{2}{*}{2nd} & 1st  & 20.63 & 5.29 & 18.25\\
      && 2nd  & 40.33 & 23.16 & 37.36\\
     \midrule
   \multirow{4}{*}{Keyword} & \multirow{2}{*}{1st} &  1st & 40.52 & 21.06 & 37.54\\
    &  &  2nd & 33.35 & 14.67 & 30.60\\
    \addlinespace[0.2em]
     \cdashline{2-6}
     \addlinespace[0.2em]
    & \multirow{2}{*}{2nd} & 1st & 22.49 & 7.26 & 20.17\\
      && 2nd & 28.75 & 12.65 & 26.19\\
     \midrule
    \multirow{4}{*}{Relation} & \multirow{2}{*}{1st} &  1st & 40.45 & 21.05 & 37.52 \\
    &  &  2nd & 33.56 & 14.65 & 30.79 \\
    \addlinespace[0.2em]
     \cdashline{2-6}
     \addlinespace[0.2em]
    & \multirow{2}{*}{2nd} & 1st & 22.85 & 7.47 & 20.47\\
      && 2nd & 28.48 & 12.42 & 25.89 \\
     \midrule
    \multirow{4}{*}{Retrieve} & \multirow{2}{*}{1st} &  1st & 39.32 & 19.74 & 36.32 \\
    &  &  2nd & 33.89 & 15.29 & 31.14\\
    \addlinespace[0.2em]
     \cdashline{2-6}
     \addlinespace[0.2em]
    & \multirow{2}{*}{2nd} & 1st & 22.61 & 7.55 & 20.34\\
      && 2nd & 28.31 & 12.33 & 25.72\\
   \bottomrule
    \end{tabular}
    \caption{ \label{tab:control_2} We divide each summary reference into two halves and deduce the oracle guidance from them separately. Feeding incompatible guidance signals can lead to degraded performance.}
  \end{table}
In addition to the qualitative results in the main paper, we also perform a quantitative analysis to demonstrate the controllability of our models.

The quantitative results in Table 3 of the main text already demonstrate to some extent that we can control the model with guidance signals, as guidance signals of better quality can lead to better summaries.
To further demonstrate this, we randomly sample guidance signals multiple times and plot the correlation between guidance quality and output quality in Figure~\ref{fig:control}. We can clearly see that there is a strong correlation between these two variables, indicating the controllability of our model.

In addition, we try to divide each test reference summary into two halves, then use oracle extraction to obtain guidance signals for both of these two halves and feed them to the model. Table~\ref{tab:control_2} shows that feeding incompatible guidance signals can lead to degraded performance, which further demonstrates that we can control the summary through provision of user-specified inputs.

\subsection{Semantic Similarity}
To evaluate the semantic similarities between our model outputs and the reference, we also compute the METEOR scores~\cite{banerjee2005meteor}. As shown in Table~\ref{tab:meteor}, all of our guided models can outperform BertAbs in temrs of both of METEOR. However, it is surprising that BertExt achieves the best performance, possibly because METEOR has a tendency to favor long summaries.
 \begin{table}[t]
       \small
  \centering
  \begin{tabular}{@{}lrrr@{}}
  \toprule
   \multirow{2}{*}{\bf Model} &  \multicolumn{2}{c}{\bf METEOR} & \bf \multirow{2}{*}{\#Words (k)} \\
   & exact match & + stem/syn/para \\
  \midrule
  BertExt &  22.24 & 20.69 & 828.62\\
  BertAbs &  19.43 & 18.01  & 669.16\\
  \midrule
  \multicolumn{2}{l}{\it Ours} \\
  \midrule
    \multirow{1}{*}{Sentence}  & 20.21 & 18.88 & 626.73\\
   \multirow{1}{*}{Keyword} &  20.16 & 18.70 & 700.48\\
    \multirow{1}{*}{Relation} &  20.12 & 18.60 & 749.30\\
    \multirow{1}{*}{Retrieve} &  19.59 & 18.07 & 717.22\\ \bottomrule
    \end{tabular}
    \caption{ \label{tab:meteor} Semantic similarity evaluation. We report results both in exact match mode (rewarding \emph{exact matches}
between words) and full mode (rewarding \emph{matching stems}, \emph{synonyms} and \emph{paraphrases} as well).}
  \end{table}
  \begin{table}[tb]
        \small
  \centering
  \begin{tabular}{@{}lrrrrr@{}}
  \toprule
 \bf Model & \bf Train &  \bf Test & \bf R-1 &\bf R-2 &\bf R-L \\
  \midrule
 \multirow{4}{*}{Sentence} & \multirow{2}{*}{Oracle} & Auto & 43.78 & 20.66 & 40.66 \\
    & & Oracle & 55.18 & 32.54 & 52.06\\
   \cdashline{2-6}
  & \multirow{2}{*}{Auto} & Auto & 41.61 & 19.04 & 38.65 \\
 &    & Oracle & 43.07 & 20.79 & 40.13\\
 \midrule
 \multirow{4}{*}{Keyword} & \multirow{2}{*}{Oracle} & Auto & 42.21 & 19.36 & 39.23\\
    & & Oracle &  45.08 & 22.22 & 42.07\\
   \cdashline{2-6}
  & \multirow{2}{*}{Auto} & Auto &  41.72 & 19.15 & 38.78\\
 &    & Oracle &  41.76 & 19.25 & 38.83 \\
 \midrule
 \multirow{4}{*}{Relation} & \multirow{2}{*}{Oracle} & Auto & 41.40 & 18.66 & 38.40 \\
    & & Oracle & 45.96 & 23.09 & 42.92\\
   \cdashline{2-6}
  & \multirow{2}{*}{Auto} & Auto &  40.29 & 18.30 & 37.33 \\
 &    & Oracle &  40.67 & 18.41 & 37.70 \\
 \midrule
 \multirow{4}{*}{Retrieve} & \multirow{2}{*}{Oracle} & Auto & 40.88 & 18.24 & 37.99\\
    & & Oracle & 43.69 & 20.53 & 40.71\\
   \cdashline{2-6}
  & \multirow{2}{*}{Auto} & Auto &  40.86 &18.5 &37.95 \\
 &    & Oracle &  41.45 & 18.86 & 38.46\\
\bottomrule
    \end{tabular}
    \caption{ \label{tab:oracle2} Using automatically constructed guidance during training degrades the performance significantly.}
  \end{table}

\begin{table*}[thb]
    \small
  \centering
    \setlength{\tabcolsep}{1.5pt}
  \begin{tabular}{@{}lrrrrrrrrrrrrrrrrrrrrr@{}}
      \toprule
      \multicolumn{1}{l}{\multirow{2}{*}{\bf Method}} &\multicolumn{6}{c}{\bf CNNDM} & & \multicolumn{6}{c}{\bf XSUM} & &  \multicolumn{6}{c}{\bf NYT} \\
      \cmidrule{2-7}\cmidrule{9-14}\cmidrule{16-21}
      & \multicolumn{3}{c}{\bf XSUM}&\multicolumn{3}{c}{\bf NYT}& & \multicolumn{3}{c}{\bf CNNDM} &\multicolumn{3}{c}{\bf NYT} & & \multicolumn{3}{c}{\bf CNNDM}& \multicolumn{3}{c}{\bf XSUM} \\
  \midrule
      BertExt  & 20.55 & 2.84 & 15.55 & 44.80 & 24.35 & 41.37 & & 35.98 & 13.38 & 32.56 & 37.35 & 16.67 & 33.84 & &  40.18 & 17.21 & 36.40 & 19.93 & 2.75 & 14.94\\
      \midrule
      BertAbs  &  20.39 & 2.85 & 15.89 &  40.99 & 20.41 & 37.91 & & 26.31 & 5.54 & 21.80 & 20.60 & 3.75 & 16.53 & & 35.77 & 14.24 & 32.67 & 16.11 & 2.24 & 12.85\\
      Ours & 20.55&2.89& 16.00 & 43.55 & 21.83 & 40.51 & & 26.72 & 5.62 & 22.08 & 23.74 & 3.61 & 18.37 & & 36.23 & 14.37 & 33.15 & 16.14 & 2.14 & 12.92\\
   \bottomrule
    \end{tabular}
    \caption{\label{tab:da} Performance of sentence-guided model under domain adaptation settings. The first row and second row represent source and target domains respectively.}
  \end{table*}

\subsection{Automatic Factual Correctness Evaluation}
\begin{figure}[t]
\centering
\includegraphics[width=0.45\textwidth]{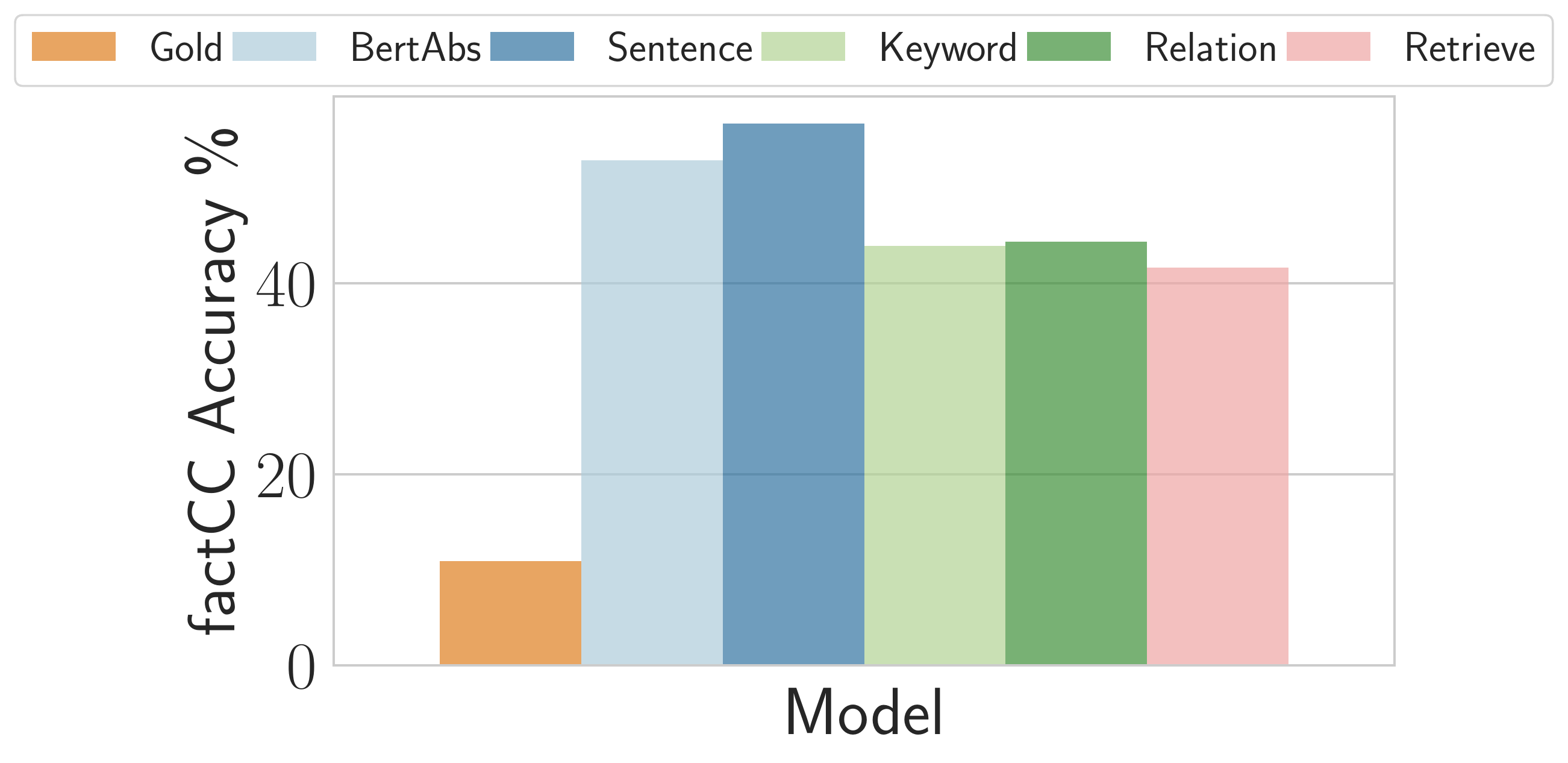}
\caption{The factCC model will give the gold reference an accuracy of about 10\%.}
\label{fig:bar_factcc}
\end{figure}

Besides human evaluation, we have also tried to use factCC~\citep{kryscinski2019evaluating}\footnote{\url{https://github.com/salesforce/factCC}} to evaluate the factual correctness of our model outputs automatically. However, as shown in Figure~\ref{fig:bar_factcc}, the factCC tool will give the gold reference an accuracy of about 10\%. Considering our model is optimized towards the gold reference, the factCC score might not be a good indicator of whether there are factual errors in a generated summary.

\subsection{Necessity of Using Oracles During Training}
We have demonstrated in the main paper that it is necessary to use an oracle to select guidance signals during training for highlighted sentence models. In this part, we investigate if this is true for all the three guidance signals as well. Table~\ref{tab:oracle2} shows that this methodology will lead to significantly worse performance for other guidance signals as well, which further verifies our hypothesis that when the relevancy between guidance and reference is weakened, the model will not learn to depend on the guidance signals and thus the model will be reduced to the original abstractive summarization baseline.

\subsection{Domain Adaptation.}

We also evaluate the performance of our highlighted sentence-guided models under domain adaptation settings, namely train a summarization model on one dataset and test it on some other datasets. As shown in Table~\ref{tab:da}, generally, extractive models can outperform abstractive ones under domain adaptations settings and our model can achieve better performance than abstractive baselines. However, while our model is given the extracted sentences by the extractive model, we still cannot outperform extractive baselines. These negative results indicate that our model may still fail to fully depend on guidance signals when doing adaptation. Possible future directions include dropping out the input documents occasionally during training so that the model can learn to better condition on the guidance.

\end{document}